\documentclass[conference,letterpaper]{IEEEtran}
\IEEEoverridecommandlockouts
\usepackage[left=0.75in,right=0.75in,top=0.75in,bottom=0.8in]{geometry}
\usepackage[utf8]{inputenc}
\usepackage{cite}
\usepackage{amsmath,amssymb,amsfonts}
\usepackage{algorithmic}
\usepackage{graphicx}
\usepackage{textcomp}
\usepackage{xcolor}
\usepackage{comment}
\usepackage{hyperref}
\usepackage{rotating}
\usepackage{multicol}
\usepackage{multirow}
\usepackage[labelformat=simple]{subcaption}
\usepackage{mathtools}
\usepackage{lipsum}
\usepackage{placeins}

\renewcommand{\IEEEtitletopspaceextra}{0.25in\relax}
\usepackage{tikz}
\usetikzlibrary{quotes,angles}
\usetikzlibrary{backgrounds}
\usetikzlibrary{positioning}
\usepackage{tkz-euclide}
\usepackage[ruled, vlined, linesnumbered]{algorithm2e}
\def\BibTeX{{\rm B\kern-.05em{\sc i\kern-.025em b}\kern-.08em
    T\kern-.1667em\lower.7ex\hbox{E}\kern-.125emX}}
    
\DeclareCaptionLabelSeparator{periodspace}{.\quad}
\title{\LARGE \bf Probabilistically Informed Robot Object Search with Multiple Regions}

\author{Matthew Collins$^{\dagger 1}$, Jared J. Beard$^{\dagger 2}$, Nicholas Ohi, Yu Gu$^{\dagger 3}$
\thanks{$^{\dagger}$Department of Mechanical, Materials, and Aerospace Engineering, West Virginia University, Morgantown, WV}
\thanks{$^{1}$ M. Collins: \texttt{mac00072@mix.wvu.edu}, $^2$ J. J. Beard: \texttt{jbeard6@mail.wvu.edu}, $^3$ Y. Gu: \texttt{yu.gu@mail.wvu.edu}}
}

\begin{document}

\maketitle

\begin{abstract}
% Big motivation (search and rescue)

% Problems with lit

% How we handle that

% Results

% The pressing need for efficient search and rescue operations in hazardous environments underscores the importance of autonomous robotic systems. Despite significant advancements, existing literature often falls short in overcoming the difficulty of long planning horizons and dealing with limited sensor information. This study introduces a novel approach that formulates the search problem as a belief Markov decision processes with options (BMDP-O) to make Monte Carlo Tree Search (MCTS) viable tools for overcoming these challenges effectively. By abstracting multi-step policies into options, our formulation enhances computational efficiency and decision-making in long-term planning. We further propose a mechanism for segmenting regions of interest (ROIs) and updating belief states, leveraging Bayesian principles to refine the search strategy in response to new observations. Our experimental results demonstrate the superiority of our approach in various scenarios, showcasing significant improvements in search efficiency and adaptability across different fields of view (FOVs).

% updated abstract
The increasing use of autonomous robot systems in hazardous environments underscores the need for efficient search and rescue operations. Despite significant advancements, existing literature on object search often falls short in overcoming the difficulty of long planning horizons and dealing with sensor limitations, such as noise. This study introduces a novel approach that formulates the search problem as a belief Markov decision processes with options (BMDP-O) to make Monte Carlo tree search (MCTS) a viable tool for overcoming these challenges in large scale environments. The proposed formulation incorporates sequences of actions (options) to move between regions of interest, enabling the algorithm to efficiently scale to large environments. This approach also enables the use of customizable fields of view, for use with multiple types of sensors. Experimental results demonstrate the superiority of this approach in large environments when compared to the problem without options and alternative tools such as receding horizon planners. Given compute time for the proposed formulation is relatively high, a further approximated ``lite" formulation is proposed. The lite formulation finds objects in a comparable number of steps with faster computation.

% \lipsum[1-2]
\end{abstract}

% search, search and rescue,, decision making, planning, regions of interest, object search, MCTS, POMDP, POMDP search, uncertainty, autonomous search, Bayesian search, Bayesian probability autonomous planning, target search, 

\section{Introduction}

Autonomous robotic systems have the potential to offer a safer and more efficient alternative to human-led searches, especially in hazardous environments. This potential is highlighted by several studies \cite{bernard2011autonomous,zuzanek2014accepted, ko2009intelligent}, which illustrate the use of autonomy in locating victims and accessing areas beyond human reach. The scope of autonomous search is not limited to emergency scenarios; it also includes everyday tasks like locating items within homes or offices \cite{zhang2019efficient, joho2011learning, aydemir2011search}. However, the utilization of autonomous systems for search faces challenges, such as managing the complexities of long-term planning with limited computational resources and dealing with uncertainties which can arise from imperfect object detection and limited fields of view.

When navigating complex environments, estimates of object positions and conditions can change over time as more observations are made. This results in the “curse of history”, which refers to the exponential growth of possible scenarios an agent must consider as it accumulates more observations over time \cite{silver2010monte}. While approaches such as greedy methods \cite{meister2020best} and receding horizon planners can mitigate inefficiencies associated with randomness \cite{Bircher2016Receding} their trajectories tend to be long and results degrade with increasing uncertainty. Methods of search which use random walks \cite{viswanathan2000levy} are especially inefficient, given they cannot use prior knowledge. Despite advances to improve long term planning such as guided sampling \cite{Kurniawati2010Motion} or hierarchical methods \cite{kim2019bi}, there is still much work to be done in the field of autonomous search. 

\begin{figure}[t!]
    \centering
    \includegraphics[width=0.98\linewidth,trim={0 100px 0 200px},clip]{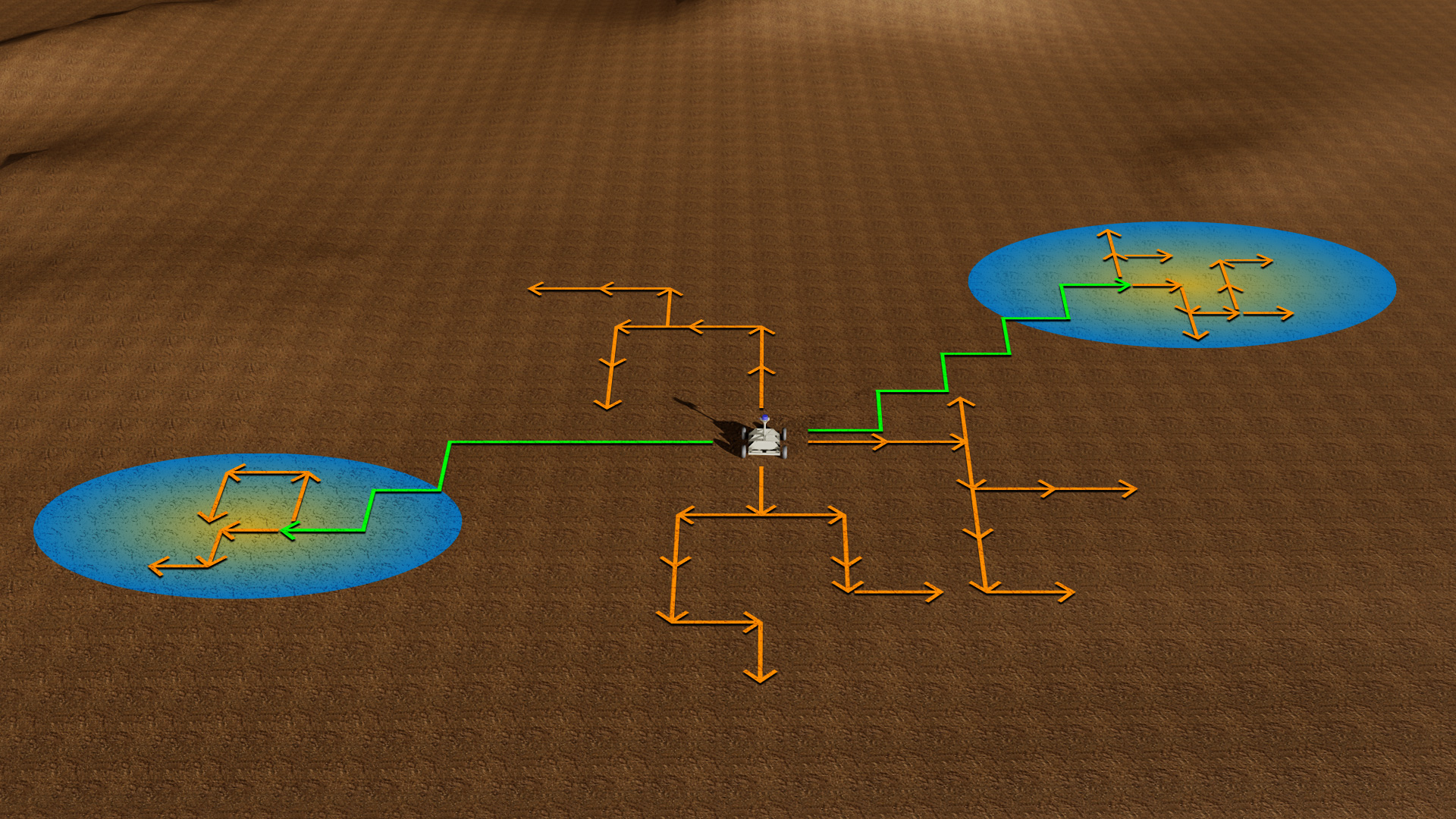}
    \caption{This figure illustrates the planning process for a robot that is searching for an object. The shaded areas represent regions of interest, having a high probability of containing the object. The robot is equipped with the proposed search method that enables it to execute single-cell movements (orange arrows) and travel to regions through a series of actions called options (green arrows).}
    \label{fig:enter-label}
\end{figure}
\FloatBarrier

Object detection in autonomous search is challenged by sensor limitations, environmental diversity, and occlusions that obscure targets. Bredeche, \textit{et al.} \cite{1048411} and Kim, \textit{et al.} \cite{Kim2011Optical} explore these challenges, highlighting issues like adaptability and processing time. Lazarevich, \textit{et al.} \cite{lazarevich2023yolobench} benchmarks and compares different state-of-the-art algorithms, and it can be seen that, while these solutions have advanced significantly, there is still a considerable amount of improvement to be made.

Our main contributions are:
\begin{itemize}
    \item A novel formulation for the search problem as a belief Markov decision process with options (BMDP-O). By abstracting motion to regions of interest (ROI) into multi-step policies, options, this formulation can balance shorter and longer horizon planning. 
    \item An approximate ``lite" formulation of the BMDP-O problem. Successive belief updates from executing options become expensive. By approximating this to an MDP-O, the decision maker can achieve similar search times but with faster computation.
    \item The proposed formulations permit generic fields of view to be used. This results in more adaptability and customization across multiple types of sensors.
\end{itemize}

The remainder of this paper is structured as follows: Section \ref{sec:prev_work} reviews related literature, Section \ref{sec:methods} explains our methodology, Section \ref{sec:results} presents and discusses the results, and Sections \ref{sec:conclusion},\ref{sec:future} conclude with a summary and future directions, respectively.

\section{Previous Work} \label{sec:prev_work}

For the purposes of our discussion, we concern ourselves with object search where the target is static. There are however, a number of variants which deal with moving or evading targets \cite{chung2011analysis}. Notably, object search is also tightly related to a number of other robotics problems such as coverage and vehicle routing \cite{otto2018optimization}, though these are special cases beyond the scope of our discussion.

Object search grew out of the Koopman's landmark works in stochastic optimization for naval applications \cite{chung2011analysis, otto2018optimization, stone1989or}. These works framed the problem from a Bayesian perspective with the focus being on analytically allocating search effort given some prior distribution \cite{koopman1957theory}. Into the late 1980s, a number of thrusts built on these results to better understand the mathematics, explore algorithmic solutions to leverage newly inexpensive computers, and later dynamically respond to changing priors \cite{stone1989or}. More recent efforts consider false detections \cite{kriheli2016optimal}, as well as multi-agent and unmanned systems across a number of domains \cite{hu2014multi}.

\subsection{Prior Information}

The primary objective of search is to find a target using the minimum effort (measured by metrics such as time, energy, or looks \cite{el2016maximum} and probability measures \cite{koopman1957theory}). Accurate prior information can drastically improve search effectiveness by guiding agents to the most relevant portions of the search space. A number of works focus on improving the quality of this prior information through means such as semantic search, often inspired by how humans search for objects \cite{zhang2019efficient, joho2011learning, aydemir2011search}. Many of these works leverage information within structured environments\textemdash those having regular or consistent, similar features. Features in structured environments can  then be used to generate informative prior distributions about objects locations. While this creates challenges of its own, structured environments, can therefore limit the number of search locations, simplifying the problem of finding minimum effort trajectories. To contrast, this work deals with unstructured environments where objects and their surroundings may not have any exploitable information between, which can lead to priors having high variance and covering large areas. 
% . Thus, as with earlier works, we assume priors are supplied as a probability distribution over the search space.

\subsection{Solutions}

Interestingly, search is a problem where the myopic or greedy strategy is the optimal solution, though this is under special conditions \cite{assaf1985optimal}. This has made greedy algorithms a desirable choice even when faced with imperfect observations \cite{kriheli2016optimal}. The entire set of conditions for optimality is still subject of study, however, with the use of more generic formulations (\textit{i.e.}, arbitrary field of view or motion constraints), the problem becomes intractable \cite{chung2011analysis}. This, in turn, has motivated the use of approximate methods such as beam search \cite{meister2020best}, mixed integer linear program solvers \cite{sato2010path}, and branch-and-bound \cite{washburn1998branch}. More recently, works have started to frame search as a POMDP \cite{you2019deep, liact, chen2024pomdp}. By framing the problem as a POMDP, powerful general purpose tools can be used to solve sequential decision making tasks. However, these methods remain computationally demanding and struggle with long planning horizons \cite{beard2020environment}. This limits their use, especially in real-time systems. 
% To combat this, a number of works exist to limit or decrease the computation required to solve POMDPs, such as \cite{roy2005finding}, \cite{holzherr2021efficient}, \cite{silver2010monte}.

% How to optimize and impacts of constraints

\subsection{Observations}

The process of search is a function of how an agent observes its environment. Whereas early works, such as that of Koopman, assumed observation of arbitrary points with certainty \cite{koopman1957theory}, works have increasingly factored in more generic sensors. A major form of uncertainty occurs through perception (\textit{e.g.}, real world systems may have false \cite{kriheli2016optimal} or ambiguous detections \cite{el2016maximum}). Alternatively, the body of the robot or environmental parameters such as sunlight may interfere with these observations \cite{gu2018robot}. Another important aspect is the field of view. Often robots have cameras, for which observation uncertainty increases with distance and orientation \cite{otto2018optimization}. Similarly, mounting a camera differently can change the effective field of view. Despite these variations, search planners generally consider observations tied to their specific hardware setup or problem of interest.

% Otto -> ref a bunch papers talking about different FOV

\section{Methodology} \label{sec:methods}

\subsection{Problem Statement} \label{sec:prob_statement}

The primary aim of this research is to improve search time in stationary target search for a single robot operating in unstructured environments. This work assumes a 2D environment, discretized into an $N$-by-$M$ grid. The robot position is fully observable; heading is not kept as a separate variable, but is assumed to follow direction of motion. The robot can move to any of the four adjacent cells with deterministic transitions. The robot has a noisy sensor with an arbitrary field of view (FOV) for observing nearby cells. It is assumed the FOV will always be oriented in the direction of motion. Each cell within the FOV has associated detection probabilities $p_{tp}$ and $p_{tn}$ for true positives and negatives, respectively. Within the environment there is a single target randomly sampled from a known prior distribution. It is assumed this prior distribution has areas of significantly higher probability than their surroundings. The agent is expected to segment out these areas into $Q$ regions of interest (ROI), which is covered in section \ref{sec:segment_ROIs}.

\begin{figure}
    \centering
    \includegraphics[width=0.98\linewidth]{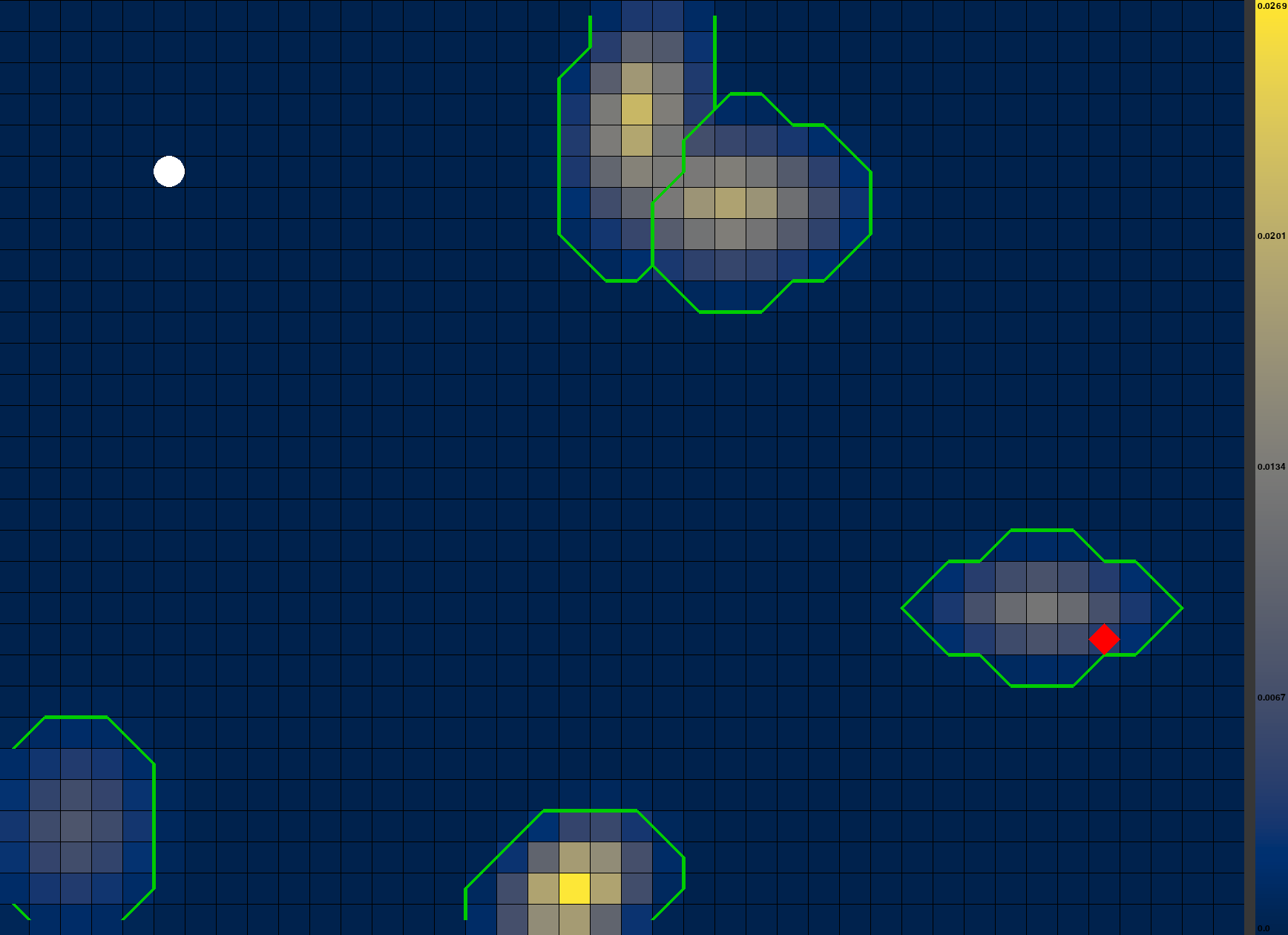}
    \caption{The robot (white circle), within the 2D grid environment. The prior probability of the object (red diamond) being in a given cell is indicated by its color. Dark blue cells represent low probability, and yellow cells represent higher probability, mapped from the colorbar to the right of the figure. Regions of interest are outlined in green.}
\end{figure}

\subsection{Problem Formulation}
\label{problem}
Framing search as a POMDP opens the door to a number of general purpose tools. Thus, POMDPs provide a convenient testing ground for understanding how the problem formulation can improve search. However, these planners must still contend with the limited FOV and long-planning horizons that make search a difficult problem. To help overcome this, we frame the problem as a finite-time belief Markov decision process with options (BMDP-O), defined as $M \coloneqq \langle \mathcal{B}, \mathcal{O}, T, R \rangle$. By using a belief MDP, we need only consider the distribution of states up to a given point, rather than the evolution of each individual state in the belief, reducing the complexity of the problem \cite{spaan2012partially}. Given there are multiple high probability regions, an agent with limited planning horizon can have difficulty navigating between these regions. Options provide a formalism to abstract a sequence of actions into a single command \cite{konidaris2007building}. This lets agents consider longer planning horizons with less computational expense than expanding a tree with each action individually.

Within the BMDP-O, $\mathcal{B}$ represents the set of belief states $b \coloneqq \langle c_{i,j},map \rangle$, where $c_{i,j} \forall i\in \{1,...N\}, j\in \{1,...M\}$ is the coordinate of the robot and $map$ is a probability distribution over potential target locations. The option set $\mathcal{O}$ contains options $o$ and actions $a \in \mathcal{A}$. Options, intended to facilitate movement about the environment, are defined as the sequence of actions to transition to neighboring ROIs. Actions include movement by one cell in any of the cardinal directions on the grid; abusing notation, let $\mathcal{A} \subset \mathcal{O}$. The transition probabilities $T(b,o,b') = p(b'|b,o)$ describe the distribution of beliefs $b'$ which result as a consequence of taking option $o$ at belief $b$. 
%This component models how the robot's actions affect its understanding and perception of the environment. This work assumes <something about transition between states>. Furthermore…. something about observations and belief updates below in Sec. {ref} 
Lastly, the reward function $R(b,o,b')$ establishes the optimization criteria with respect to a given transition. Let $m_c \in map$ be the mass in a given cell, and $d$ be a time penalty. The reward is defined as 
\begin{equation}
    R(b,o,b') = 
    \begin{cases}
        \sum_{c\in FOV}m_c - d & \text{if } o \in \mathcal{A}\\
        \sum_{c \in \mathcal{T}}\left(\sum_{c\in FOV}m_c - d\right), & \text{otherwise}.
    \end{cases}
\end{equation}
In the first case, assuming the agent takes an action, the robot receives a reward equal to the mass within the FOV, less some time discount. Otherwise, the agent sums up the mass in all FOVs observed during the trajectory $\mathcal{T}$ implied by option $o$.

% is defined in terms of the belief probabilities for each cell in the discretized 2D grid. Specifically, R calculates the reward for a robot's path as the cumulative belief probabilities of the cells it traverses. Formally, for a path P comprising a set of cells, and b(c) denoting the belief probability for cell c, the reward is R(P) = SUM(b(c) for all P in c). This design motivates the robot to navigate towards areas with higher probabilities of object locations, aligning with its updated belief state and optimizing search efficiency.

\subsection{Segmenting ROIs}
    \label{sec:segment_ROIs}
Segmenting the prior map into ROI constrains the decision problem to high probability cells. These ROI, however, are relatively sparse. Decision making tools such as MCTS struggle to plan over long distances, meaning it can be difficult to navigate to ROI in highly discretized environments. Path planners like A\textsuperscript{*} \cite{hart1968formal}, on the other hand, are well suited to this task. Thus, segmenting the map into ROI lets an agent leverage domain knowledge in path planning to more effectively navigate these large swaths of low-value cells. The decision maker can then focus search on the areas of high probability. Given there are no obstacles, we can simplify the planning task further. Rather than solving the path explicitly, the option begins by specifying the most probable cell in the specified ROI. Then the option greedily selects actions until the agent arrives at the desired cell.

Segmentation was conducted prior to search. While methods such as clustering \cite{saxena2017review} can be used, testing found clustering to produce more inconsistent segmentation. Instead, the watershed method was used \cite{watershed}. First local maxima are identified to find the number of ROI $Q$. These maxima act as seeds from which the watershed method grows the ROI. Neighboring cells are incorporated into the region, continuing this expansion until it encounters the boundary of an adjacent region or the probability of neighboring cells fall below the user-defined threshold $\tau$. 
% This systematic and iterative inclusion of cells ensures comprehensive coverage, resulting in each cell being assigned to a specific region or being excluded based on the threshold criterion. This method effectively highlights the most promising areas for object discovery, optimizing the search strategy in a focused and efficient manner.

\subsection{Belief Update} \label{sec:belief_update}

Since multiple grid cells can be observed simultaneously from one observation, the measurement is $Z = \{z_{i,j}: \forall c_{i,j} \in FOV\}$. The measurements $z_{i,j}$ may be either positive $z^+$ or negative $z^-$ with probabilities $p_{tp}$ and $p_{tn}$, as described in Sec. \ref{problem}.
% This information can be tracked solely in the marginal probabilities of the other ROIs, because the belief state of each grid cell is conditionally independent, and the marginal probability of an ROI can be found as the sum of the probabilities of all grid cells it contains, as in \eqref{eq:roi_marginal_prob}.
By the Law of Total Probability, the probability of an observation of grid cell $z_{i,j} \in Z$  is
\begin{equation}
    \label{eq:evidence}
    P(z_{i,j}) = P(z_{i,j} | z^\pm)P(z^\pm) + \\
    P(z_{i,j} | z^\mp)P(z^\mp)\text{.}
\end{equation}
The posterior probability the object is in cell $c_{i,j}$ can be found using Bayes' Rule:
\begin{equation}
    \label{eq:cells_in_m_update}
    P(c_{i,j} | z_{i,j}) = \frac{P(z_{i,j} | c_{i,j})P(c_{i,j})}{P(z_{i,j})}\text{.}
\end{equation}

In practice, computing this update over all cells can be quite expensive. This is especially true in tree search where such an update may be applied several hundred times. As an approximation, the distribution is instead updated as follows for each step in the tree search. We take advantage of $P(z_{i,j})$ being a normalization constant applied to all terms and let
\begin{equation}
    P(c_{i,j}|z_{i,j}) \approx P(z_{i,j} | c_{i,j})P(c_{i,j})\text{.}
\end{equation}
From there, we make two simplifying assumptions. First, the search is sparse so the agent will rarely encounter an object. Thus we assume all observations will be negative. Should a positive observation occur, it would show up in the prior at the next step. Next, we assume the field of view is much smaller than the environment. As such, changes in probability of observed regions will be much greater than the mass redistributed over the remaining area. Additionally, because planning horizon will be limited relative to the scope of the problem, cumulative errors will be limited in most cases. We can therefore reduce computational expense by only to updating only the cells observed in $Z$, as follows 
\begin{equation}
    P(c_{i,j}|z^-) = P(z^-|c_{i,j})P(c_{i,j}) = (1-p_{tp})P(c_{i,j})
\end{equation}
for cells in the FOV. Here the second equality follows from the requirement that $P(z^-|c_{i,j})+P(z^+|c_{i,j}) = 1$.

\subsection{Lite Formulation}

Despite the speed improvements from using options, belief updates can become burdensome in large environments. This motivated the use of a further approximated\textemdash lite\textemdash formulation for the problem. As such, belief updates are not calculated and the prior map is assumed to remain static. Thus the problem formulation is reduced to an MDP-O, $\mathcal{M}\coloneqq \langle S,O,T,R\rangle$. Now, the states are simply the agent coordinates $s = c_{i,j}$ and transitions $T(s,o,s')$ are only focused on the motion of the agent to a different cell. We now have the updated reward function
\begin{equation}
    R(s,o,s') = 
    \begin{cases}
        \sum_{c\in FOV}m_c - d & \text{if } o \in \mathcal{A}\\
        \dfrac{f\Bar{m}_{ROI_k}}{A^2d_{ROI_k}}, & \text{otherwise}.
    \end{cases}
\end{equation}
Here $f$ is a tuning parameter, $\Bar{m}_{ROI_k}$ is the mean mass in ROI\textsubscript{k}, $A$ is the area of the map, and $d_{ROI_k}$ is the distance from the centroid of ROI\textsubscript{k}. For options, this formulation returns the anticipated density of searching an ROI rather than the observed mass (as in the BMDP-O). 

% \subsection{Algorithms}

% The search methods included were MCTS, MCTS with regions, MCTS with regions lite, receding horizon search, and breadth-first search. MCTS with regions was directly compared to MCTS to evaluate the impact of region considerations when planning. Receding horizon search was chosen for its proficiency in managing uncertainties or noise within the environment. Breadth-first search served as a baseline for comparison against a more straightforward, greedy approach. The inclusion of MCTS regions lite was to provide a less computationally intensive alternative to MCTS with regions, as stated in the previous section. 

\section{Results} \label{sec:results}

This section presents the results obtained from simulations that allow a single robot search for an object in a 200-by-200 environment described in Section \ref{sec:prob_statement}. Recall that in the full model, the agent can only take a single action; options are reserved for use by the select planners. For each scenario, 250 trials were conducted.

To generate the priors, $N\sim unif(1,5)$ Gaussians were sample with means uniformly distributed throughout the environment and variance $\sigma^2\sim unif(0.04,0.12)$ in both dimensions. Convolving these distributions gave the prior, from which the object location was sampled. After each time step update was computed for use as the prior in the next time step. Note, this belief update did not rely on the approximations in Section \ref{sec:belief_update} as these were only used for planning.

\subsection{Baseline Comparisons}

Testing centered on two thrusts: controlling for the algorithm employed and for the improvements of the proposed formulations. Each algorithm assumes a point observation with $p_{tp}=p_{tn}=0.9$. Results are reported as 5\textsuperscript{th}, 25\textsuperscript{th}, 50\textsuperscript{th}, 75\textsuperscript{th}, and 95\textsuperscript{th} percentiles to avoid null solutions, such as the robot spawning next to the object. 

As baselines, we select two receding horizon approaches, given their demonstrated use in search. First is a greedy planner, which moves the agent towards the highest valued cell within the given horizon. Limiting the horizon reduced the computational burden associated with convolving the FOV with the entire search space. The second planner was a form of direct policy search (DPS). DPS works better than value search in some high dimensional problems \cite{kochenderfer2015decision}, making it a suitable candidate for search problems having large maps. The DPS planner employed an $\epsilon$-greedy rollout with $\epsilon=0.65$. Both receding horizon planners were given horizons of 20 steps. These algorithms were compared against predictor upper confidence trees (PUCT) \cite{rosin2011multi}, a commonly used variant of MCTS. PUCT was allocated a budget of 40 iterations, rollout of 60 steps, and exploration parameter $c=0.5$. All three algorithms were given the simulator described above\textemdash not considering ROIs and options in their decision making. 

As demonstrated in Figure. \ref{fig:fov_steps}, PUCT provided the most consistent, best search times making it the prime candidate from which to test the proposed formulations. Thus by comparing baseline simulator with the BMDP-O (PUCT Regions) and Lite MDP-O (PUCT Regions Lite) formulations, both their search performance and the computational expense could be compared. For the PUCT Regions Lite formulation, the tuning parameter $f$ was set to $8*10^{-6}$. 

 % MCTS with regions was directly compared to MCTS to evaluate the impact of region considerations when planning. The inclusion of MCTS regions lite was to provide a less computationally intensive alternative to MCTS with regions, as stated in the previous section.

% \subsection{Baseline Comparisons}

% rename receding horizon to Direct Policy Search
% rename bfs to greedy search
% rename MCTS to PUCT
\begin{figure}
    \centering
    \includegraphics[width=0.98\linewidth]{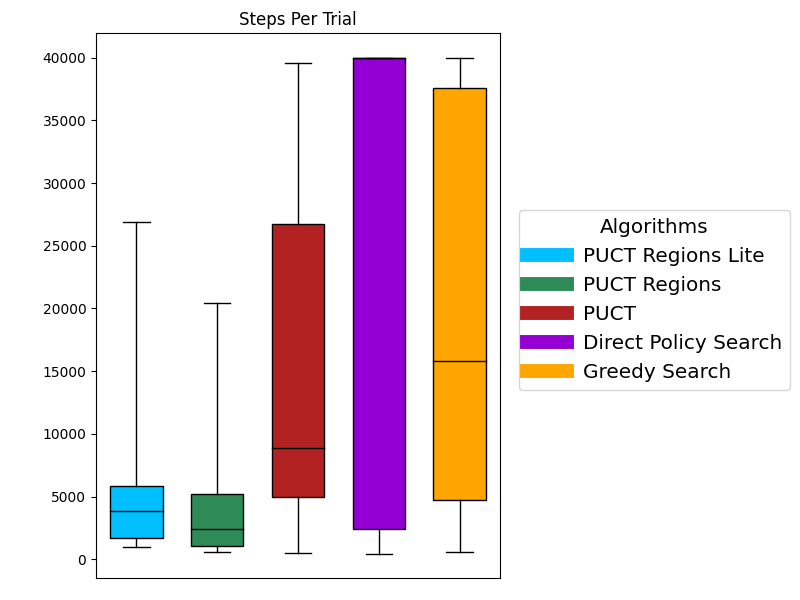}
    \caption{Number of steps taken to find the target object per trial within a 200-by-200 grid environment. Each bar is reported at the 5\textsuperscript{th}, 25\textsuperscript{th}, median, 75\textsuperscript{th}, and 95\textsuperscript{th} percentiles. PUCT consistently outperforms DPS and greedy search. PUCT Regions Lite and PUCT Regions have comparable search times, with most trials doing better than the full model without options.}
    \label{fig:alg_data}
\end{figure}

% Each search method was gauged through two primary metrics: the number of steps per experiment and the time taken to evaluate a solution at each step. These metrics have seen common use historically \cite{} give the importance of time in applications such as search and rescue missions.

\begin{table}[h!]
    \centering
    \begin{tabular}{cccccc}
        \hline
        Percentiles & 5\textsuperscript{th} & 25\textsuperscript{th} & Median & 75\textsuperscript{th} & 95\textsuperscript{th} \\
        \hline
        PUCT Regions Lite & 23.38 & 27.92 & 38.11 & 47.97 & 59.10 \\
        PUCT Regions & 41.82 & 56.87 & 66.55 & 75.50 & 87.06 \\
        PUCT & 11.56 & 18.67 & 18.80 & 19.84 & 19.91 \\
        Direct Policy Search & 40.49 & 51.82 & 53.93 & 58.58 & 59.04 \\
        Greedy Search & 24.35 & 28.05 & 33.19 & 37.63 & 41.02 \\
        \hline
    \end{tabular}
    \caption{Time Per Step [ms]}
    \label{tab:performance_percentiles}
\end{table}

This work is intended to solve search in real time, making step time integral to efficient execution of search. 
On median, DPS and greedy search were $\sim$1.8 and $\sim$2.9 times slower, respectively (Table \ref{tab:performance_percentiles}). 
Furthermore, over 50\% of DPS trials exceeded the maximum number of steps allocated (Fig. \ref{fig:alg_data}). 
While greedy search performed better, its median result was still $\sim$6934 steps longer than that of PUCT. 
Coupled with their relatively long searches (Fig. \ref{fig:alg_data}), they were deemed unsuitable for further testing.

% This efficiency gain offered by PUCT Regions Lite is particularly advantageous in real-world search scenarios, since the time spent traveling or maneuvering is often a substantial portion of the total operation time.

% Comparing the median times per step within Table \ref{tab:performance_percentiles}, PUCT Regions Lite performs similarly to greedy search, and it takes about twice as long per step as PUCT. PUCT Regions takes about the same amount of time per step as DPS, but finds objects much faster, as seen in Figure \ref{fig:alg_data}.

Within Table \ref{tab:performance_percentiles}, our findings revealed that PUCT Regions Lite significantly reduced the median time per step by $57$\% compared to PUCT Regions, albeit at the cost of an increased number of steps per trial by $61$\% (Fig. \ref{fig:alg_data}). Much like DPS and greedy search, PUCT Regions Lite and PUCT Regions were nearly 2 and 3 times slower than PUCT, respectively.
Looking back to the number of search steps to find the object, underscores the trade-off between computational load and operational efficiency. Despite their slower compute times, $\sim$75\% of PUCT Regions and PUCT Regions Lite trials found the object in fewer steps than the best $\sim25\%$ of PUCT trials. For both PUCT Regions and PUCT Regions lite, we conjecture the long tails of the distribution are objects which occurred outside ROI. PUCT Regions and PUCT Regions Lite covered only 5.96\% and 9.71\% of the total area on median. Based on these results, PUCT Regions Lite is best on resource constrained platforms, such as drones, and platforms with near real-time constraints. Alternatively, in scenarios where efficiency is more important PUCT Regions offers shorter search times, both in median and with tighter variance.

\subsection{Fields of view}

\begin{figure}
    \centering
    \includegraphics[width=0.3\linewidth]{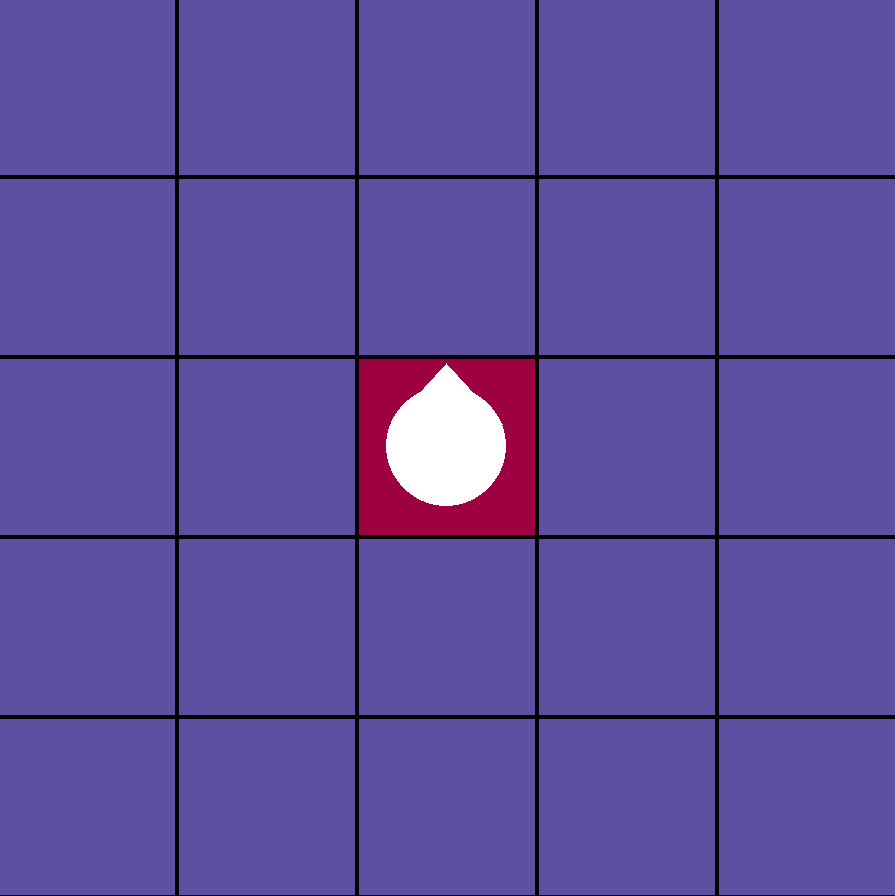}
    \includegraphics[width=0.3\linewidth]{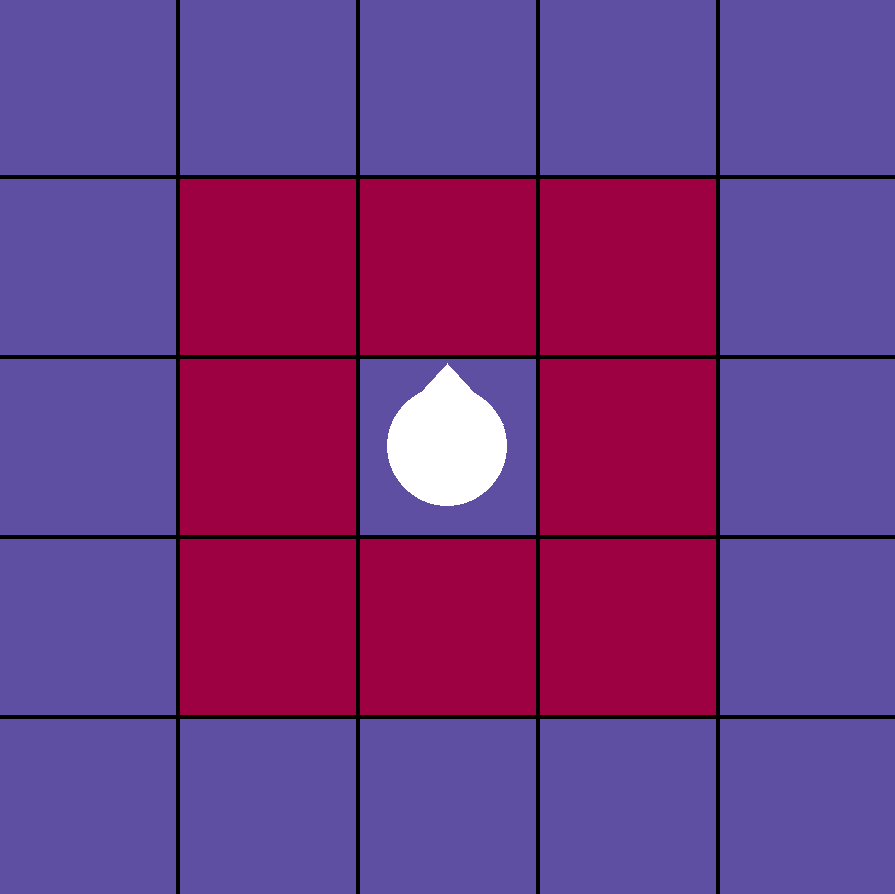}
    \includegraphics[width=0.3\linewidth]{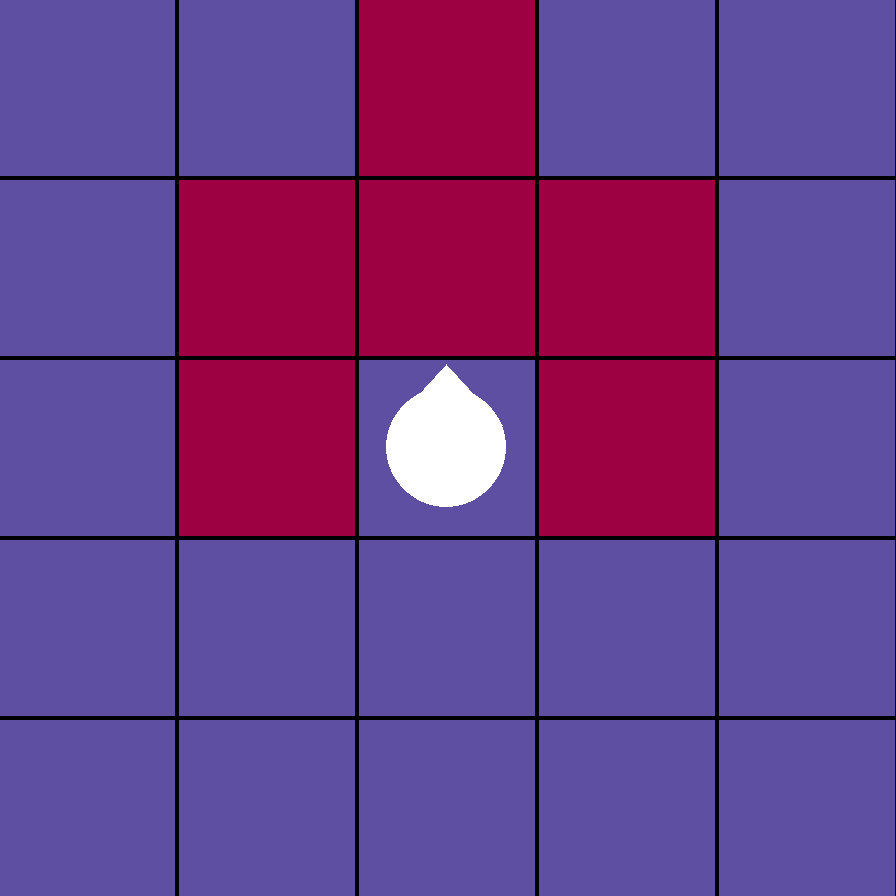}
    \caption{The different fields of view used within simulations. With the robot being in the center cell and facing up, the red cells are observable, and the blue cells are unobservable. From left to right, the FOVs are denoted as point, donut, and forward-facing wide-angle camera observations}
    \label{fig:fovs}
\end{figure}
% single cell fov: point observation
% donut fov: donut fov, robot blocks the middle cell
% front facing fov: forward facing camera/wide angle camera
To better understand the capabilities of the PUCT Regions Lite formulation and test its compatibility with multiple platforms, multiple FOV were considered (Fig. \ref{fig:fovs}). Point observations were selected to model highly abstracted problems where an agent could check its immediate surroundings. Two other FOV were selected to model cameras. The donut-shaped FOV represents a downward facing camera with the robot body occluding the center, as on the Cataglyphis rover \cite{gu2018robot}. The forward-facing camera models a wide angle lens oriented in the direction of motion. For simplicity, it is a assumed all cells within FOV have $p_{tp}=p_{tn}=0.9$.

\begin{figure}
    \centering
    \includegraphics[width=0.98\linewidth]{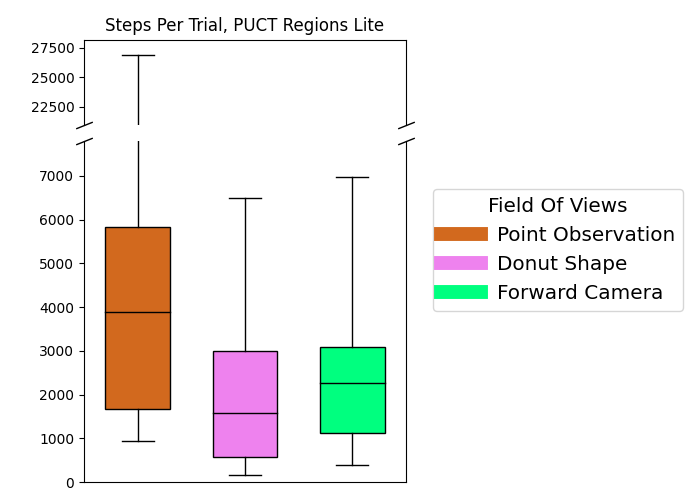}
    \caption{Number of steps taken to find the target object per trial within a 200-by-200 grid environment. Each bar represents the different fields of view when used with MCTS regions lite. The corresponding fields of view can be seen in Fig. \ref{fig:fovs}. With an increasing number of observable cells, search time decreases, though with diminishing returns.}
    \label{fig:fov_steps}
\end{figure}

As seen in Fig. \ref{fig:fov_steps}, a point observation tends to need more steps per trial to find an object versus multi-cell FOVs, since there are less cells being observed per time step. Of note, the multi-cell FOV have a $95^{th}$ percentile $\sim25\%$ of the point observation. Comparing the multi-cell FOVs, the donut-shaped FOV performs slightly better than the forward-facing camera FOV. This can be attributed again to the capability of observing more cells. Interestingly, the forward facing camera has a tighter distribution, indicating the asymmetry made be advantageous to improving consistency, though this would require further study to verify.

% \subsection{Limitations}

% long trails likely objects outside ROI....

\section{Conclusion} \label{sec:conclusion}

In this paper, a novel formulation of the search problem has been demonstrated to leverage standard decision making tools. The formulation uses motion planning solutions as options. This abstracts movement to ROI, limiting time spent searching low probability areas.
%and a belief MDP to be less cumbersome while allowing the use of POMDP tools. 
The simulation results demonstrated an increase in computational efficiency and a decrease in search time when compared to other search methods in large-scale environments. We further demonstrate an approximate formulation which retains many of the search efficiency gains but at a reduced computational load.
%This increased efficiency when searching in large environments is attributed to the ability to travel to different regions, enabling a more global view of the process, instead of only searching locally.
% Furthermore, in comparison with existing approaches, our method has shown to efficiently search large environments with multiple valuable regions. 

% Existing methods, such as receding horizon and MCTS, are able to search a single region well, but they are unable to consistently travel to other valuable regions due to their limited planning horizon. Our method is able to search nearby as well, but with the addition of considering each region when planning, both a local and a more global view of the environment can be more efficiently searched in less time, due to the robot spending more time in valuable areas, and less time in lower valued areas. Our method is also extendable to arbitrary FOVs, allowing for more wide spread use and adaptation.

% new paragraph version from above
Given their limited planning horizons, formulations without options are unable to consistently travel to valuable regions due to their limited planning horizon. With the addition of considering regions when planning, use of options enables our method to make both long and short horizon decisions, increasing the depth of search. This leads to more time spent searching in high valued areas. 

Our method is also extendable to arbitrary FOVs, allowing for more wide spread use and adaptation. Equipping different sensors, the number of steps taken tends to lessen when working with larger FOVs, due to more area being observed at a time.

\section{Future Work} \label{sec:future}

Despite these advancements, it is also important to acknowledge the limitations of this work. Notably, the current approach is designed for discretized environments, which may not fully capture the complexities of continuous spaces. Future research in continuous environments could 
%enhancing its applicability across a broader spectrum of search scenarios and 
allow for more accurate modeling of the robot's motion and observations. Additionally, in vast environments with numerous regions, the method's planning process may become inefficient, as it attempts to calculate paths and rewards for each region at every step. A promising direction for future work could involve constraining the total number of regions considered for planning at each step or use of methods such as progressive widening to limit tree expansion \cite{sunberg2018online}. Moreover, integrating multi-resolution or hierarchical frameworks could further address computational challenges, offering a more scalable solution for extensive search operations. Another limitation is the assumption of static environments; future efforts could aim to develop methodologies that account for multiple agents, dynamically updating ROI, and evading targets. 
% Exploring the extension of our methodology to multi-agent systems also represents a significant avenue for future research.

% In conclusion, this work helps enhance the problem of search in large environments by efficiently using the information given to it, which advances the field of autonomous search to allow for more use and the adoption of autonomous robots.

\section*{Acknowledgment}

This research was made possible by Alpha Foundation, and the NASA West Virginia Space Grant Consortium, Grant \#80NSSC20M0055.

\bibliographystyle{IEEEtran}
\bibliography{root}

\end{document}